\documentclass[11pt,a4paper]{article}
\usepackage[hyperref]{acl2019}
\usepackage{times}
\usepackage{latexsym}
\usepackage{mathtools}
\usepackage{url}
\usepackage{booktabs}
\usepackage{caption,subcaption}

\aclfinalcopy 

\title{Learning to Predict Novel Noun-Noun Compounds}

\author{Prajit Dhar \\
  Leiden University \\
  \texttt{dharp@liacs.leidenuniv.nl} \\\And
  Lonneke van der Plas \\
  University of Malta \\
  \texttt{lonneke.vanderplas@um.edu.mt}}

\date{}

\begin{document}
\maketitle
\begin{abstract}
We introduce temporally and contextually-aware models for the novel task of predicting unseen but plausible concepts, as conveyed by noun-noun compounds in a time-stamped corpus. We train compositional models on observed compounds, more specifically the composed distributed representations of their constituents across a time-stamped corpus, while giving it corrupted instances (where head or modifier are replaced by a random constituent) as negative evidence. The model captures generalisations over this data and learns what combinations give rise to plausible compounds and which ones do not. After training, we query the model for the plausibility of automatically generated novel combinations and verify whether the classifications are accurate. For our best model, we find that in around 85\% of the cases, the novel compounds generated are attested in previously unseen data. An additional estimated 5\% are plausible despite not being attested in the recent corpus, based on judgments from independent human raters.
\end{abstract}

\section{Introduction}

Compounding is defined as the process of combining two or more lexemes to form a new concept \cite{bauer2017}. For most compounds in English, the first constituent is the modifier, whereas the second is the head. The head usually determines the class to which the compound belongs, whereas the modifier adds specialisation, e.g \textit{apple cake} is a type of \textit{cake}. Compounding is thought of as one of the simplest forms of concept formation\footnote{We avoid the usage of ``word formation'' due to there being no consensus on the definitions for both words and compounds \cite[chapter 2]{bauer2017}.} as it involves use of elements that are already part of the language and requires little or no morphological changes, particularly in English. From the perspective of language acquisition, \citet{ruthy} found that children acquired compounding construction skills before the other forms of word formation. 

Comparatively little effort has been put into investigating the productive word formation process of compounding computationally.  
Although compounding is a rather challenging process to model as it involves concepts of compositionality and plausibility along with an intricate blend of semantic and syntactic processes, it is, in our view, one of the best starting points for modeling linguistic creativity. In contrast to relatively more studied topics in linguistics creativity, such as automatic poetry generation \cite{ghazvininejad-etal-2017-hafez}, aesthetics are not involved. Moreover, compounding is limited to phrase level processes, as it involves a combination of known lexemes.

In general, the creative power of language has been understudied in the field of natural language processing (NLP). The main focus is indeed on processing, as the name suggests. Creative thinking is a cognitive ability that fuels innovation. Therefore, the modelling and understanding of the underlying processes of novel concept creation is relevant. Ultimately, we aim to create tools that enhance people’s ability to interface more creatively with large data sets, to build tools that find inspiration in data.

Our main contributions are the introduction of a new task in NLP that sheds light on the basic mechanisms underlying conceptual creativity; an automatic way of evaluating newly generated language; a temporally-aware neural model that learns what are plausible new conceptual combinations by generalising over attested combinations and corrupted instances thereof.

\section{Related Work}

The related task of automatic novel compound detection was introduced by Lapata and Lascarides~\shortcite{LapataLascarides}. Their aim is to distinguish rare noun compounds from rare but nonce noun sequences. The biggest difference between their work and ours is that while they identify existing, albeit rare, and therefore possibly relatively novel compounds in corpora, we predict unseen, and therefore novel compounds, in an absolute sense. Still, the overlap between the tasks makes the work relevant. In their experiments, surface features, such as the frequency of the compound head/modifier, the likelihood of a word as a head/modifier, or the surface-grammatical context surrounding a candidate compound perform almost as well as features that are estimated on the basis of existing taxonomies such as WordNet. Although the semantic features they gathered from WordNet did not do very well, we believe our distributional semantic features are more fine-grained. The simple statistical features that did well in distinguishing rare compounds from nonce terms, would not be suitable in our scenario, where we try to generate novel, plausible compounds. We did however, follow their methodology for the automatic extraction of noun-noun compounds from corpora based on their PoS. 

\citet{KellerLapata2003} obtain frequencies for unseen bigrams (including noun-noun bigrams) in corpora using Web search engines and show evidence of the reliability of the web counts for natural language processing, also by means of studying the correlation between the web counts and human plausibility judgments. The unseen bigrams are generated in a random fashion from corpus data, as the aim is not to generate plausible combinations, but to overcome data sparseness by providing counts from Web searches. 

\citet{o-seaghdha-2010-latent} uses topic models for selectional preference induction, and evaluates his models on the same data as \citet{KellerLapata2003} outperforming previous work. As this work tries to predict the plausibility of unseen combinations, it is more closely related to our work. We are, however, first and foremost interested in the temporal aspect of novel compound creation, and therefore use a time-stamped corpus and temporally-aware models. We also use this time-stamped corpus for evaluation, in addition to human plausibility judgments.

We find a number of works in the related field of cognitive science that focus on predicting human acceptability or plausibility ratings with compositional distributional models.
\citet{journals/cogsci/VecchiMZB17} focus on novel adjective noun-phrases. They show that the extent to which an adjective alters the distributional representation of a noun it modifies is the most significant factor in determining the acceptability of the phrase.  \citet{GuentherMarelli} are also concerned with predicting plausibility ratings, but focus on noun-noun compounds instead. The main difference between their work and ours is the fact that our systems are partly neural and use slightly different features, and aim to generate novel, plausible compounds that are evaluated by checking for their existence in a future corpus, whereas they check for correlation with human plausibility ratings on a set of attested and corrupted compounds. However, their careful investigation of the different distributional semantic factors in their model have been very insightful for us and they inspired one of our systems. For example, they found that a higher relatedness between head and compound is associated with higher plausibility. And the similarity between the constituents is associated with a slightly higher plausibility as well. We used these features in one of our models as well.

More recently, \citet{Marellietal2017} presented a data-driven computational system for compound semantic processing based on compositional distributional semantics and evaluate the model against behavioral results concerning the processing of novel compounds. They find that the phenomena of relational priming and relational dominance are captured by the compositional distributional semantic models (cDSMs), whose predictions pattern nicely with the results from the behavioral experiments. Although this work proves that the cDSM is psychologically real when it comes to processing novel compounds, and we find inspiration in the architecture of their model, their work is mainly aimed at modelling compound processing, whereas we are focusing on compound prediction.

For the architecture of our model, we were mainly inspired by Van de Cruys~\shortcite{Cruys14}. The problem we are focusing on has many similarities with the task that their paper focuses on: predicting selectional preferences for unseen data. We adopted the architecture of the neural network from this paper as well as the method to generate negative data for training purposes, originally proposed by \citet{Collobert}. Apart from the difference in the task we are trying to address, the main differences between their work and ours is the fact that we are adding a temporal aspect to the neural networks. 

\section{Novel Compound Prediction}

In this paper, we address the task of novel compound prediction. Three models are created that use count-based distributed word representations of known compound constituents to predict unseen, but plausible compounds.

\subsection{Intuitions and evaluation}

In particular, we address the task of predicting novel noun-noun (N-N) compounds: compounds consisting of two constituents that are both nouns. Our method is based on the generalisation power of machine learning. We reason that by compressing the space in ways that are in line with distributional patterns found in observed data, estimates for unobserved yet plausible combinations should be close to the estimates gathered from attested examples. For example, if we have seen glass-bottom boats in corpora, but we have never seen the combination \textit{glass canoe} (a recent invention), we can infer from the similarity between the components of the compounds that a glass canoe could be a plausible compound even though it has never been seen.

Evaluating plausibility prediction models for novel combinations is non-trivial and previous work has relied mainly on human judgments \cite{GuentherMarelli}. We aim to find an automatic evaluation method to ease parameter optimisation and model comparison. To this end, we use a time-stamped corpus divided into decades. The last decade is used for testing, with the previous decades used for training the models. This allows us to check whether the novel generated compounds are good predictions of compounds that might emerge in the future. Because the future extends beyond the last decade of our corpus, and the results from our automatic evaluation are pessimistic, we ask human judges to rate the plausibility of a sample of automatically generated novel compounds that are not attested in the last decade (see Sections \ref{system_eval} and \ref{hum-eval}). 

\subsection{Two Aspects for Compounding}
We hypothesise that in order to computationally model the phenomenon of compounding, we need our models to be both contextually-aware and temporally-aware, which we explain in detail in the subsequent section.

\subsubsection{The Contextual Aspect}

Psycholinguistic research on N-N compounds in Dutch seems to suggest that constituents such as -molen `-mill' in pepermolen `peppermill' are separately stored as abstract combinatorial structures rather than processed on-line and understood on the basis of their independent constituents: molen `mill' \cite{dejongetal2002}. For English open compounds, a similar effect was found for the right constituent. To test if this phenomenon has an effect in the process of generating novel compounds, we decided to experiment with two types of contextual contexts: \textbf{CompoundCentric} and  \textbf{CompoundAgnostic}. \\

\begin{figure*}
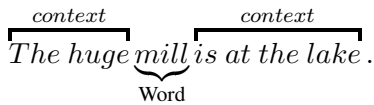
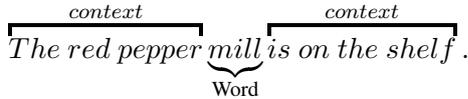


\centering
\begin{subfigure}[b]{0.45\linewidth}
\textbf{Head context} \\ 

$\overbracket{The\ red\ \underbrace{pepper}_\text{Modifier}}^{context} \underbrace{mill}_\text{Head} \overbracket{is\ on\ the\ shelf}^{context}.$ \\

\textbf{Modifier context} \\

$\overbracket{The\ new}^{context}\underbrace{mill}_\text{Modifier}$ $\overbracket{\underbrace{stone}_\text{Head}\ has\ started}^{context}.$ 
\caption{\label{df}}
\end{subfigure}
\begin{subfigure}[b]{0.45\linewidth}
$ \overbracket{The\ huge}^{context} \underbrace{mill}_\text{Word} \overbracket{is\ at\ the\ lake}^{context}.$

$\overbracket{The\ red\ pepper}^{context} \underbrace{mill}_\text{Word} \overbracket{is\ on\ the\ shelf}^{context}.$

$\overbracket{The\ new}^{context}\underbrace{mill}_\text{Word}$ $\overbracket{stone\ has\ started}^{context}.$

\caption{\label{sf}}
\end{subfigure}

\caption{Contexts for (\subref{df}) CompoundCentric and (\subref{sf}) CompoundAgnostic aspects}
\end{figure*}

\textbf{CompoundAgnostic}: These are the standard window-based contexts used in vector-based representations of words. We capture the distributional vectors of the words, irrespective of whether lexemes are found as constituents of a compound or as simple standalone words (see Figure \ref{sf}).

\textbf{CompoundCentric}: To the best of our knowledge, distributional models that are sensitive to the role a lexeme plays in a compound have not been tested before. Here we capture the distributional vectors of words based  on their usage as constituents in a compound. So the word \textit{mill} would have different representations, depending on its role in a compound. In Figure \ref{df}, we show an example context that \textit{mill} gets as a head, and an example context it gets as a modifier.


\subsubsection{The Temporal Aspect}

Previous works such as \citet{sploosh} have shown that meanings of certain words change over time. The same can be observed for compounds such as \textit{melting pot}. The meaning of \textit{melting pot} deviated from its original meaning ("A blast furnace") to its current meaning, that of a society where people from many different cultures are living together, and assimilating into a cohesive whole. To test if time does impact our task, we envision two settings for our models:

\textbf{DecadeCentric}: In this setting, we emphasise the temporal aspect by collecting counts of individual compounds and their constituents per decade. We reason that knowing about the usage trend of the constituents of a compound might help to predict which constituents will be combined next. For example, if a certain word is trending, you would expect it to crop up in novel combinations.

\textbf{DecadeAgnostic}: To test if our intuition about the temporal aspect indeed holds true, we also collect the counts of the individual compounds and their constituents without any temporal information.

\section{System Overview}

\begin{figure}
    \centering
    \includegraphics[width=\linewidth]{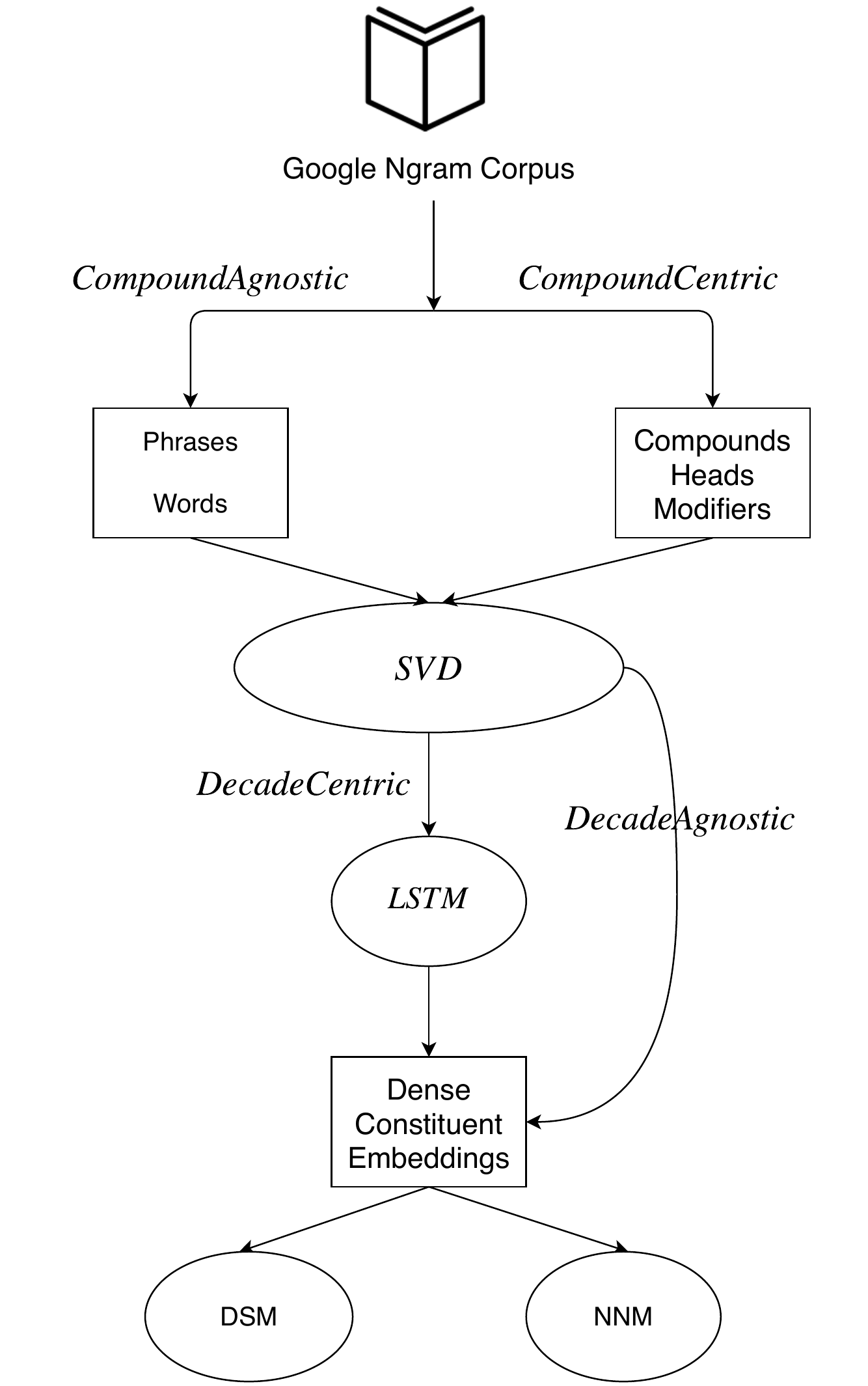}
    \caption{System Overview for the DSM and NNM models}
    \label{fig:system_overview}
\end{figure}

Figure \ref{fig:system_overview} shows the system overview for our two main models. The two aspects explained in the previous section are clearly visible as distinct routes in the system overview. From the Google Ngram Corpus \cite{googlengrams}, distributional vectors that are either CompoundAgnostic or CompoundCentric are collected\footnote{Even though the CompoundAgnostic aspect returns phrases and words instead of actual compounds and constituents, respectively, we refer to them as same for simplicity.}. The counts are either collected per decade (DecadeCentric) or without any temporal information (DecadeAgnostic). The vectors then undergo dimensionality reduction (to 300 dimensions) using singular-value decomposition (SVD), after which they are either directly input to our two semantic models (DecadeAgnostic), or passed through a Long short-term memory (LSTM) model (DecadeCentric) before they are input to the two semantic models. The reason for doing so is that it makes it easier to compare the models for each aspect. The LSTM we use takes a sequence of constituents representations for each decade as input and returns a single representation for the constituent that ideally encompasses its entire history.
In the following subsections, we will provide more details on the data processing we performed.

\section{Data} 

We will describe how we collected our data, pre-processed it and how we generated negative data for the classifier.

\subsection{Data Collection}
We constructed four different datasets based on the aforementioned aspects from the Google Books Ngram corpus \cite{googlengrams}. This corpus describes the usage of words and phrases over a period of five centuries. Information of around 6\% of all books in eight languages was gathered using Optical Character Recognition (OCR). For our study, we focused on the extraction of English unigrams and bigrams from the corpus and on 20 decades (1800s to 1990s), with the final decade (2000s) only used to collect the newly generated compounds. The corpus was tagged using a Conditional Random Field (CRF) based tagger \cite{Lafferty}, which has an error rate of 1-5\%. The universal PoS tagset \cite{postagset} was used to perform the PoS tagging.

\subsection{Data Pre-processing}

Similar to \citet{LapataLascarides}, we made sure that our compounds are not surrounded by nouns, in order to avoid extracting parts of compounds that have more than two constituents. Also, constituents and compounds containing non-alphabetic characters are excluded from the experiments. For each compound and its constituents, a context window of 3 words is used (i.e. 3 words on either side of the target word), in order to retrieve their distributional representations. We were initially limited to a 2-word context window size due to the nature of the Google Ngram 5-gram corpus. However with the use of a sliding window approach, this was increased to a window size of 3. A bigram (say the compound \textit{water cycle}) could occur in four different positions in the 5-grams (1-2, 2-3, 3-4 and finally 4-5). The contexts for each of these positions is then captured.

We only consider the top 50k most commonly occurring nouns, adjectives, verbs and adverbs to be candidate contexts. Finally, the heads are lemmatised and converted to lowercase. A compound is considered to be novel if it only exists in the final decade (the 2000s) and has a frequency count of at least 3. We chose the cut off count of 3, so as to capture most rare and plausible compounds but at the same time eliminate hapax legomena (terms that occur only once).


\subsection{Negative Data Generation}

In order to train the classifiers for the task of predicting plausible novel compounds, we need both positive and negative data. The positive class for validation and testing is comprised of the modifier and head of compounds that were newly created in the decades 1990s and 2000s, respectively. 
In the absence of an attested negative class, i.e. compounds that are implausible, we used the strategy from \citet{Collobert} to generate our own negative class. This class is made up of corrupt tuples, that are constructed by randomly replacing one of the constituents in the tuple ($m$,$h$) with a corresponding constituent (heads are replaced by heads and modifiers by modifiers) from another attested compound.

We then have two scenarios: the \textbf{CorruptHead} and the \textbf{CorruptModifier} scenarios. For a \textbf{CorruptHead} scenario, the head $h$ is replaced by randomly selected head $h'$ with the modifier remaining the same. 
Similarly, for a \textbf{CorruptModifier} scenario, we replace the modifier $m$ with $m'$ (see Table \ref{disamhead}).


\paragraph{Disambiguation Task} The models have the task of disambiguating the attested tuples from their corrupted counterparts. The purpose of allowing these two scenarios is to test whether corrupting the head leads to better negative data points, than corrupting the modifier. We also make sure that none of the corrupted tuples from the aforementioned procedures results in the generation of a novel or a previously existing compound.

\begin{table*}
\begin{center}
\begin{tabular}{cccccccc}
\toprule
\multicolumn{2}{c}{CorruptHead} & & \multicolumn{2}{c}{Novel Compound} & & \multicolumn{2}{c}{CorruptModifier} \\
$m$ & $h'$ & & $m$ & $h$ &  & $m'$ & $h$ \\
\midrule
water & fox && water & absorption &  &blanket & absorption\\
pitch & minister & $\leftarrow$ & pitch & accent & $\rightarrow$ & cement & accent\\
gene & psychiatry & & gene & sequence & & dolphin & sequence  \\
\bottomrule
\end{tabular}
\end{center}
\caption{Generation of negative examples using the CorruptHead and CorruptModifier scenarios}
\label{disamhead}
\end{table*}

\section{Semantic Models}

In Figure \ref{fig:system_overview}, we show that the distributional vectors, be they CompoundAgnostic or not and be they DecadeAgnostic or not are fed into two distinct semantic models. Note that at this point the distributional vectors are turned into dense representations. We experimented with one additional model, the Distributional Feature Model (DFM), that uses sparse embeddings for constructing its features. The other two models:  the Distributional Semantic Models (DSM) and Neural Network Model (NNM) use dense embeddings. \footnote{All the implementation details are provided in the GitHub repository \url{https://github.com/prajitdhar/Compounding.}}

\subsection{Distributional Feature Model}
In the DFM, for each compound, distributional semantic features are constructed. The raw frequency counts were used to construct the features as weighting measures such as PPMI worsened the final results. The first three are adopted from Information Theory and commonly used to find collocations between words (for more detail see \citet{Manning1999}) -
\begin{enumerate}
\item Positive Pointwise Mutual Information \textit{PPMI}: A variation of the Pointwise Mutual Information (\textit{PMI}) where the negative \textit{PMI} values are replaced by 0's. \textit{PPMI} is preferred over \textit{PMI} as it has been shown to outperform PMI on semantic similarity tasks \cite{bullbull}.  The  \textit{PPMI} for a compound $comp$ and its two constituents $m$ and $h$ is defined as
\begin{equation}
\textit{PPMI}(comp) = max(\log_2 \frac{P(comp)}{P(m)P(h)},0),
\end{equation}
where $P(comp)$ is the probability of both $m$ and $h$ occurring together (i.e. the compound itself).

\item Log likelihood-ratio \textit{LLR}: \textit{PPMI} scores are biased towards rare collocations as they assign rare words with rather high \textit{PMI} values. To overcome this bias and to incorporate the frequency counts of the constituents, the log likelihood ratio is used as another feature, similar to \citet{llr}.

\item Local Mutual Information  \textit{LMI}: It is another metric that tries to overcome the bias of \textit{PPMI} and does so by comparing the probability of observing $m$ and $h$ (or $comp$) together with the probability of observing the two by chance:

\begin{equation}
\begin{split}
\textit{LMI}(comp) & = P(comp) \cdot \log_2 \frac{P(comp)}{P(m)P(h)} \\
& = P(comp) \cdot PMI
\end{split}
\end{equation}

The three features below relate to the calculation of the similarity between the three components of the compound. The similarity between any two target words is defined by the cosine similarity between their vectors.

\item Similarity between the Compound and its Constituents \textit{sim-with-head} and \textit{sim-with-mod}: The similarity between a compound $comp$ and a constituent $c$ is defined as:

\begin{equation}
cos(\overrightarrow{comp},\overrightarrow{c}) = \frac{\overrightarrow{comp} \cdot \overrightarrow{c}}{\| \overrightarrow{comp} \| \| \overrightarrow{c} \|},
\end{equation}
where $\overrightarrow{comp}$ and $\overrightarrow{c}$ are the vector representation of $comp$ and $c$, respectively.


\item Similarity between the Constituents: The similarity between the constituents of a compound is computed as well. \citet{GuentherMarelli} and \citet{lynott} have found this score to be useful in discerning plausible compounds. Formally, the similarity between a modifier $mod$ and a head $head$ is defined as:

\begin{equation}
cos(\overrightarrow{mod},\overrightarrow{head}) = \frac{\overrightarrow{mod} \cdot \overrightarrow{head}}{\| \overrightarrow{mod} \| \| \overrightarrow{head} \|},
\end{equation}
where $\overrightarrow{mod}$ and $\overrightarrow{head}$ are the vector representation of $mod$ and $head$, respectively.

\end{enumerate}
Finally to retrieve the features for the constituents, the distributional features of the compounds are averaged. We also collect the standard deviation of each feature, so as to get the best approximation of the distribution of the original values.
The 6 constituent features (and 114 for the DecadeCentric aspect as there are 19 decades in training) are then concatenated to represent the compound.  
So in total, we have 12 features (and 228 for the DecadeCentric aspect) as input to the DFM, which uses the stochastic gradient boosting model \cite[XGBoost,][]{Chen:2016} to perform the supervised learning. The tree-based models were trained using logistic regression as the loss function. The following parameters were then tuned and set as follows : learning rate = 0.1, maximum tree depth = 3, number of estimators = 100, minimum child weight = 6, sub-sample ratio = 0.5, $\gamma$ = 0, $\alpha$ = 0.05 and $\beta$ = 1.

\subsection{Neural Network Model}
As shown by \citet{Cruys14} and \citet{Tsubaki}, neural networks have been successful in composing a representation of a phrase or sentence. Since we expect our novel compounds to be compositional in nature, i.e. their representation could be derived from their constituents' representations, 
a compositional neural network should be able to discriminate between plausible candidate compounds and their nonsensical counterparts. 

A candidate compound is represented by its modifier and head:
\begin{equation}
x = [ i_m, j_h ],
\end{equation}
where $i$ and $j$ are the vector representations of the modifier $m$ and head $h$, respectively, and the resultant composed vector $x$ serves as the input to the neural network.The vector $x$ is then the input to the NNM.

\begin{figure}
\includegraphics[width=\linewidth]{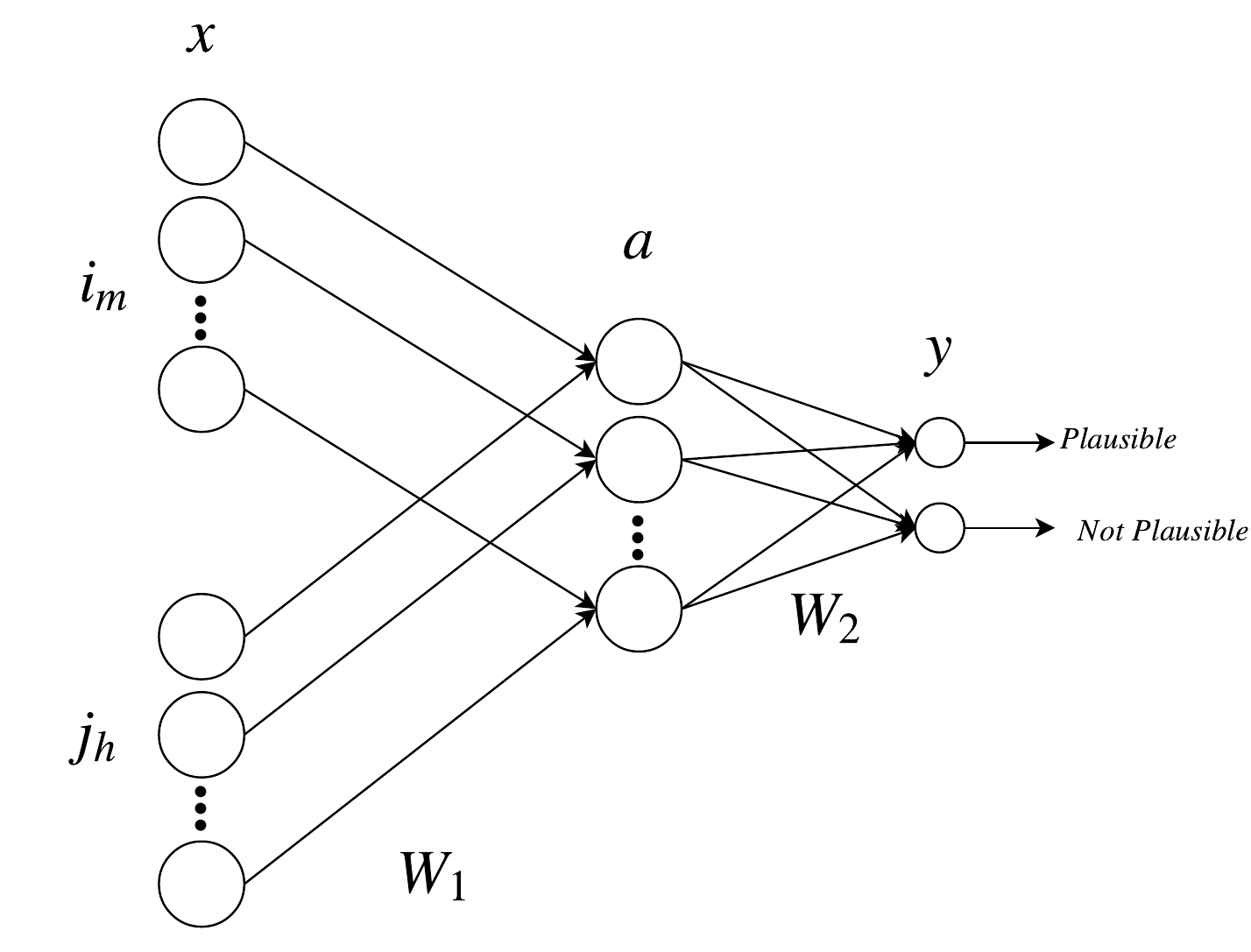}
\caption{Architecture of NNM}
\label{neuralnet}
\end{figure}

The architecture of the NNM is similar to the two-way selectional preferences model from Van der Cruys~\shortcite{Cruys14} and it comprises of a feed-forward neural network with one hidden layer as shown in Figure \ref{neuralnet}.

An element-wise activation function rectified linear unit ($ReLU$) is used in the hidden layer:
\begin{equation}
a= f(W_1 x + b_1),
\end{equation}
where $a$ is the activation function of the hidden layer with $H$ nodes, $f(.)$ performs an element-wise $ReLU$ function and $W_1$ and $b_1$ are the weights and bias of the first layer, respectively.
In the end, the NNM generates a plausibility score $y$:
\begin{equation}
y = W_2 ~a,
\end{equation}
where $W_2$ is the weight matrix for the final layer. The neural network was trained for 50 epochs with stochastic gradient descent (SGD) used for optimization. A batch size of 100 was chosen and the hidden layer was set to $H=300$ nodes.

\subsection{Distributional Semantic Models}

In order to discern if a compositional neural network is indeed required for our disambiguation task, we also implement Distributional Semantic Models (DSMs) as a baseline. Since both models (DSMs and NNM) would be using the same constituent embeddings for all four aspects, the final results would help us answer this question.
Similar to DFM, we concatenate the constituent features (here embeddings) which are then taken as input to a gradient boosting model, to predict the plausibility.

\section{System Evaluation}
\label{system_eval}
We evaluate our overall system in two ways: on the basis of corpus data, and by means of human judges, which we cover in Section \ref{hum-eval}.

For the automatic evaluation of our system, we measure how many of the compounds it predicts are attested among the previously unseen compounds in the last decade, the 2000s. The 18 decades prior to the 1990s (1800s to 1980s) are used to train the models, with the decade 1990s used for validation. The constituents that exist only in the decade of the 2000s are therefore excluded from the training phase. This way we make sure that the prediction of novel entities is dependent only on information derived from the prior decades.

\section{Results on Corpus Data}
Since all the datasets were equally balanced, we only report the accuracy scores for each of the models. We control the randomness of the negative data generation and run all our semantic models using 10 different datasets. Each of these datasets consists of the same attested novel compounds, along with a different negative class. 
The results of all the models in our experiments are shown in Table \ref{tab:results}. 

The NNM produces by far the best results. It attains an accuracy of 85\% when the CompoundCentric and DecadeCentric aspects are observed and the heads are corrupted (CorruptHead) in order to generate negative data. This means that for 85\% of the compounds generated, the model is able to correctly classify the compound as plausible or implausible. The accuracy in this case is solely determined by the fact that the novel compound is found in the last decade (that is excluded from training). We will show in a separate evaluation (see Section \ref{hum-eval}) with human judges that this score, although already quite impressive is pessimistic, because some of the novel compounds predicted by the model are not found in the last decade, but still plausible, according to human judges.

Furthermore, the following observations can be made :

Overall, the models that are sensitive to the temporal aspect (DecadeCentric), outperform models that do not take the temporal aspect into account (DecadeAgnostic). This underlines our hypothesis that the temporal aspect is crucial when modelling the process of compounding. 

Also, our hypothesis that distributional models should be sensitive to the fact that lexemes as part of a compound differ in meaning from lexemes that appear as standalone words seems to hold true. In general, models that observe the CompoundCentric aspect, perform better.
Lastly, corrupting the head to generate negative evidence seems a better alternative than corrupting the modifier. This is to be expected as the head determines most of the compound's meaning. Replacing the head with another head has a higher probability of generating an implausible compound than replacing the modifier. Implausible compounds are needed to generate negative evidence.

\begin{table*}[ht]
\centering
\begin{tabular}{ccccc}
\toprule
& \multicolumn{4}{c}{DecadeCentric} \\
& \multicolumn{2}{c}{CompoundCentric} & \multicolumn{2}{c}{CompoundAgnostic}\\
Model	&	CorruptHead &	CorruptMod 	&	CorruptHead &	CorruptMod\\
\midrule
DFM & $71.57 \pm 0.31$ & $68.76 \pm 0.2$& $70.95 \pm 0.35$ & $69.29 \pm 0.36$\\
DSM & $68.18 \pm 0.33$ & $64.77 \pm 0.26$ & $67.05 \pm 0.71$ & $64.03 \pm 0.35$\\
NNM & $84.69 \pm 0.33$ & $84.55 \pm 0.46$ & $74.32 \pm 0.63$ & $76.48 \pm 0.56$\\
\bottomrule

& \multicolumn{4}{c}{DecadeAgnostic} \\
& \multicolumn{2}{c}{CompoundCentric} & \multicolumn{2}{c}{CompoundAgnostic}\\
Model	&	CorruptHead &	CorruptMod 	&	CorruptHead &	CorruptMod\\
\midrule
DFM & $69.17 \pm 0.24$ & $66.69 \pm 0.25$ & $69.67 \pm 0.39$& $67.33 \pm 0.27$ \\
DSM & $68.26 \pm 0.43$ & $65.04 \pm 0.34$ & $67.52 \pm 0.58$ & $65.04 \pm 0.34$\\
NNM & $82.92 \pm 0.2$ & $82.54 \pm 0.4$ & $72.38 \pm 0.92$ & $75.02 \pm 0.57$\\
\bottomrule
\end{tabular}
\caption{Results of the Semantic Models, represented with accuracy and the standard deviation}
\label{tab:results}
\end{table*}

\begin{table}[h]
\centering
\begin{tabular}{ll}
\toprule
Compound  & Plausibility rating\\
\midrule

Service ramp &  4\\
Art direction & 3.34\\
Resource  companion &  2\\
Funeral fish&  0\\
\bottomrule
\end{tabular}
\caption{Human evaluations (average plausibility ratings) for compounds that are non-attested in corpus}
\label{tab:human}
\end{table}
\section{Human Evaluation}
\label{hum-eval}
Evaluation on corpus data does not guarantee full coverage. In other words, if a novel compound generated by the system is not found in the contemporary corpus this does not mean per se that the compound is not plausible. The compound  might be plausible but not yet 'invented'. We therefore also ran a small-scale manual annotation. 

Taking our best model, which was NNM under CompoundCentric and DecadeCentric aspects, a subset of plausible compounds that were predicted by our system (but not found in test corpus and hence counted as incorrect) were annotated by human judges. Following the annotation guidelines of \citet{Graves2013}, each annotator was asked to rate each candidate compound between 0 (makes no sense) and 4 (makes complete sense). 250 plausible compounds were annotated in total.

Each candidate compound was evaluated by at least three annotaters. Table \ref{tab:human} shows some of the annotation results. We see that compounds such as \textit{art direction} and \textit{service ramp}, that are predicted by the system, but not found in the decades 2000s, is deemed plausible by the annotators. In fact, we found that around $5\%$ of the test data set was rated 3 or higher, on average, by the annotators, indicating that we cannot just rely on a corpora for the evaluation of the novel compound predictor, and that the accuracies given in Table \ref{tab:results} are pessimistic.

\section{Conclusions and Future Work}
We propose a method for the task of novel compound prediction. We show that this task can be modeled computationally, and that our models need to be both temporally and contextually aware, in order to properly model compounding. The evaluation method we proposed that uses a contemporary corpus to evaluate the novel compound predicted, provides an objective and cheap alternative to evaluation with human judges. In a separate evaluation, we show that the latter provide more optimistic results. 

Although previous work has shown correlations between human plausibility judgments on unseen bigrams and frequencies in larger corpora \cite{KellerLapata2003}, we would like to study the correlation between human plausibility judgments and occurrence in the last decade, to rigorously test the viability of the evaluation method. We would also like to take the graded nature of the plausibility judgments into account when evaluating our models.

In addition, we would like to experiment with other models such as cDSMs and other neural network architectures. Our current system uses a rather simple LSTM architecture to encode temporal information into one representation, and prior tests have shown that enhanced neural network architectures such as \citet{Schuster1997BidirectionalRN} and \citet{RaffelE15} that use bidirectional LSTMS and attention-based networks, respectively, are better at encoding this information. 

Lastly, we would like to cover a more diverse set of compounds in future work. Our experiments are currently  limited to collecting N-N compounds. In subsequent experiments, we aim to add closed compounds and compounds separated by a hyphen, as well as compounds that are composed of other parts of speech, such as adjective-noun compounds.

\section*{Acknowledgements}

We would like to thank the anonymous reviewers for their valuable comments and also thank Janis Pagel for providing feedback towards the later stages of this research. 

\bibliography{novel_compounds_workshop}
\bibliographystyle{acl_natbib}

\end{document}